\documentclass[10pt,twocolumn,letterpaper]{article}

\usepackage{wacv}              

\usepackage[pagebackref,breaklinks,colorlinks]{hyperref}
\newcommand{\cmark}{\textcolor{green!70!black}{\checkmark}}
\newcommand{\xmark}{\textcolor{red!80!black}{\texttimes}}
\usepackage{tcolorbox}
\usepackage{graphicx}
\usepackage{amsmath}
\usepackage[accsupp]{axessibility}
\usepackage{pifont}
\usepackage{amssymb}
\usepackage{booktabs}
\usepackage{algorithm}        
\usepackage{algpseudocode} 
\usepackage{multirow}

\usepackage[table]{xcolor}
\usepackage{booktabs}
\usepackage{multirow}
\usepackage{adjustbox}
\usepackage{xcolor}

\usepackage[capitalize]{cleveref}
\crefname{section}{Sec.}{Secs.}
\Crefname{section}{Section}{Sections}
\Crefname{table}{Table}{Tables}
\crefname{table}{Tab.}{Tabs.}
\usepackage{subcaption} 

\definecolor{wacvblue}{rgb}{0.21,0.49,0.74}

\def\wacvPaperID{3400} 
\def\confName{WACV}
\def\confYear{2026}

\title{Beyond Real Weights: Hypercomplex Representations for Stable Quantization}

\author{Jawad Ibn Ahad\textsuperscript{1}  Maisha Rahman\textsuperscript{1} Amrijit Biswas\textsuperscript{1} Muhammad Rafsan Kabir\textsuperscript{1} 
Robin Krambroeckers\textsuperscript{1} \\ Sifat Momen\textsuperscript{2} Nabeel Mohammed\textsuperscript{2}  Shafin Rahman\textsuperscript{2}\\
\textsuperscript{1}Artificial Intelligence Department, RobotBulls Labs, Geneva, Switzerland\\
\textsuperscript{2}Machine Intelligence Lab (MILab), North South University, Bangladesh\\
{\tt\small \{jawad, maisha, amrijit, muhammad, robin\}@robotbulls.com}\\ \tt\small \{sifat.momen, nabeel.mohammed, shafin.rahman\}@northsouth.edu}

\begin{document}
\maketitle

\begin{abstract}
Multimodal language models (MLLMs) demand immense parameter capacity to align high-dimensional visual features with linguistic representations, making them computationally heavy and difficult to deploy efficiently. We introduce a progressive reparameterization strategy that \emph{physically compresses} these models by progressively replacing dense feed-forward network blocks with compact Parameterized Hypercomplex Multiplication (PHM) layers. A residual interpolation schedule, together with lightweight reconstruction and knowledge distillation losses, ensures that the PHM modules inherit the functional behavior of their dense counterparts throughout training. This transition yields substantial parameter and FLOP reductions while preserving strong multimodal alignment, leading to faster inference without degrading output quality. We evaluate the approach on vision–language models (VLMs). Our approach preserves strong performance comparable with base VLMs, while delivering substantial reductions in model size and inference latency. Progressive PHM substitution thus offers an architecture-compatible path toward more efficient multimodal reasoning and complements existing low-bit quantization techniques. Codes are available at \url{https://github.com/milab-nsu/PHM}. 
\end{abstract}


\section{Introduction}
\begin{figure}[!t]
    \centering
    \includegraphics[width=0.47\textwidth]{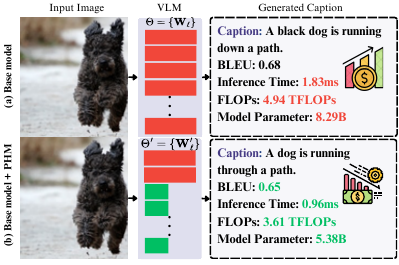} 
    \caption{Comparison of Qwen2.5-VL 7B outputs: (a) original base model vs. (b) base model with our PHM representations. The PHM variant achieves lower FLOPs, faster inference, and reduced parameters while maintaining comparable captioning quality.}
    \label{fig:fig1}
\end{figure}

Multimodal large language models (MLLMs) have become foundational to modern AI, enabling the joint processing of text and vision within transformer architectures \cite{Tsimpoukelli2021,Zeng2023}. By learning cross-modal representations, these models support a wide range of tasks, such as visually grounded dialogue, visual question answering, and multimodal retrieval \cite{Gadre2023}. Beyond academic benchmarks, MLLMs increasingly power real-world applications such as medical image–report alignment in healthcare, interactive tutors in education, and instruction-following systems in robotics \cite{Xu2022,Vasudevan2023}. Yet, state-of-the-art models typically contain billions of parameters, demanding immense computational resources for training and deployment \cite{OpenAI2023GPT4,Zhao2023Survey}. While Multimodal Optimization methods such as parameter-efficient tuning (PEFT) reduce fine-tuning overhead, they rely on adapters or low-rank modules and leave the physical parameter count unchanged \cite{Dettmers2023QLoRA,He2022UnifiedPETuning}. This gap motivates our work: designing MLLMs whose intrinsic parameterization is compact while maintaining high semantic fidelity. As illustrated in Fig.~\ref{fig:fig1}, our proposed Parameterization of Hypercomplex Multiplication (PHM) approach, which reparameterizes VLMs' dense feed-forward layers via Kronecker products of fixed hypercomplex bases with learnable compact matrices, achieves lower inference latency, reduced FLOPs, and fewer parameters than its dense baseline, while producing comparable visual captions.

Existing research on MLLM efficiency largely falls into:  
\emph{(a)} \textbf{PEFT-based:} Adapters, low-rank projections, and prompt-based tuning efficiently adapt large backbones \cite{Hu2021LoRA, Mahabadi2021Compacter, He2022UnifiedPETuning}, with further improvements from QLoRA \cite{Dettmers2023QLoRA} and multitask adapters \cite{Asai2022Multitask}. However, these methods reduce training cost, not model size.
\emph{(b)} \textbf{Structural efficiency:} Sparse attention, mixture-of-experts routing, and modular architectures improve scalability \cite{Fedus2022Switch, Clark2022UnifiedSparsity, Artetxe2021EfficientLMs}, but are not yet deeply integrated into multimodal pipelines.  
\emph{(c)} \textbf{Knowledge distillation (KD):} KD compresses large teachers into smaller students \cite{Jin2023SurveyKD, Hsieh2023DistillLM}, including multimodal variants \cite{Zhou2023MMKD, Fang2023CrossModal}. Yet, aligning dense multimodal supervision with significantly smaller models remains challenging. Overall, prior work typically addresses only one dimension, efficient adaptation, structural compactness, or knowledge transfer, leading to trade-offs in scalability, robustness, and representation fidelity, leaving open how to unify them.

To address this gap, we present a unified multimodal optimization approach that integrates PHM reparameterization, PEFT, and KD. Our framework applies LoRA to text-side attention projections, replaces feed-forward layers with residual hypercomplex blocks under convex blending, adds a lightweight reconstruction loss to stabilize dense–hypercomplex transitions, and employs a dense-teacher vs. hypercomplex-student KD strategy. Together, these enable a smooth shift from dense to hypercomplex representations, without compromising performance.

\noindent \textbf{Contributions:} \textbf{(1)} A unified framework combining hypercomplex reparameterization, LoRA adaptation, and KD for optimal and scalable multimodal learning; \textbf{(2)} Stabilization mechanisms, convex-blended hypercomplex blocks and reconstruction-based alignment, that extend recent compact multimodal designs; \textbf{(3)} Extensive experiments showing superior efficiency–performance trade-offs, where our dense-teacher vs. hypercomplex-student KD strategy preserves semantic fidelity with reduced computational overhead.

\section{Related Work}
\textbf{Hypercomplex Parameterization:}
Early work on quaternion-valued neural networks \cite{Parcollet18,Gaudet18} showed that extending weights into hypercomplex domains yields compact representations and structured weight sharing. Subsequent octonion and generalized formulations \cite{Zhu20,Zhang21Survey,Zhang21PHM} broadened model capacity and enabled scalable architectures, motivating applications in transformers and vision models \cite{Mao23,Gao23}. Residual and hybrid designs \cite{Cheng24,Wang22,Kim23} further demonstrated stability and training benefits, while recent studies highlight hypercomplexity as a promising direction for large language models and multimodal alignment \cite{Zhang24HCNN,Lee24Hybrid}. Building on this trajectory, our work employs residual hypercomplex blocks to balance dense pathways, improving efficiency and robustness.

\noindent \textbf{Knowledge Distillation:}  
Introduced by Hinton et al. \cite{Hinton15}, Knowledge Distillation (KD) compresses large teachers into smaller students by transferring softened predictions. FitNets \cite{Romero15} extended KD with intermediate supervision, while attention transfer \cite{Zagoruyko17} and relational KD \cite{Park19} exploited structural cues. Patient KD \cite{Sun19}, TinyBERT \cite{Jiao20}, and DistilBERT \cite{Sanh19} showed strong results for language models. In multimodal systems, KD has proven effective for aligning vision and language \cite{Gupta22,Li23,Wu24}. Advanced strategies include adaptive schedules \cite{Mirzadeh20,Shen20} and multi-teacher setups \cite{You17,Yang21}, enabling flexibility across tasks. Recent works extend KD to reinforcement learning \cite{Wang23RLKD}, vision–language pretraining \cite{Tang23}, and cross-modal retrieval \cite{Chen24XKD}. Building on this, we adopt a hybrid setup where a dense teacher supervises a hypercomplex student, ensuring progressive transfer and semantic alignment.

\noindent \textbf{Multimodal Optimization:}
Lightweight adaptation methods have become central to efficient multimodal fine-tuning. PEFT \cite{Houlsby19} introduced small trainable modules that preserve the base model, while LoRA \cite{Hu21} improved efficiency via low-rank updates. Subsequent variants: AdaLoRA \cite{Zhang22AdaLoRA}, DoRA \cite{Liu23DoRA}, and MixLoRA \cite{Chen24MixLoRA} refined this direction through dynamic rank allocation, decomposed updates, and expert routing. Additional strategies such as prefix-tuning \cite{Li21}, prompt-tuning \cite{Lester21}, QLoRA \cite{Dettmers24}, and recent multimodal or memory-efficient extensions \cite{Zhang23,Chen23,Mukherjee23,He24Unified,Ma24} further reduce training overhead. Quantization-aware approaches like MQuant \cite{yu2025mquant} and MBQ \cite{li2025mbq} decrease bit-precision across modalities, but mainly target post-training or fine-tuning rather than compressing the model’s underlying structure. Our work addresses this gap by combining LoRA on text-side projections with hypercomplex reparameterization of dense FFNs, stabilized through reconstruction and teacher–student distillation.

\begin{figure*}[!t]
    \centering
    \includegraphics[width=\textwidth]{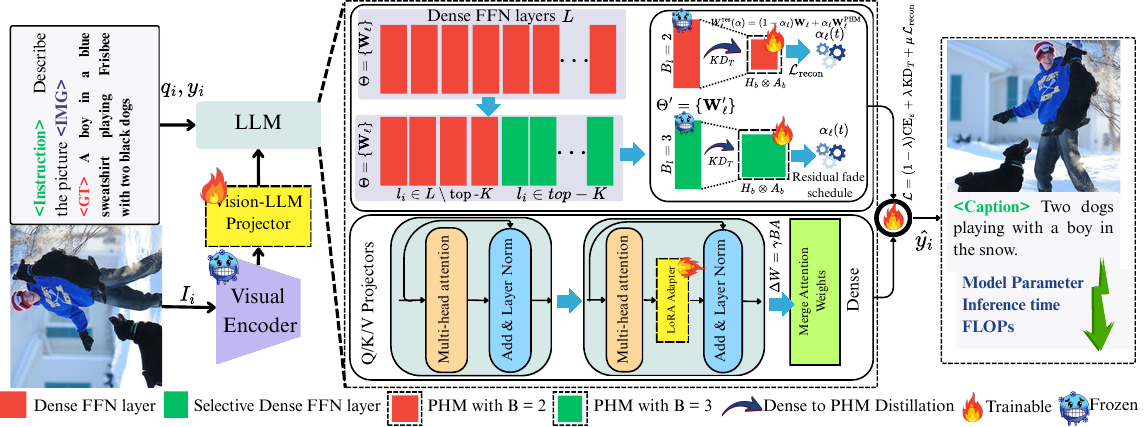} 
    \caption{Given an image and text prompt, the image is encoded by a frozen vision encoder and converted into visual tokens, which are combined with text tokens and passed into the transformer. Attention blocks operate normally with light LoRA tuning, while the heavy FFN blocks are progressively replaced by compact PHM layers. Each FFN is reparameterized as a residual mix of its original dense weight and a PHM operator built from a few fixed $2\times2$ bases $H_b$ and small learnable core matrices $A_b$. Early in training, the dense FFN performs most computation, and PHM contributes little, but through a smooth fade-in schedule guided by Knowledge Distillation and a reconstruction loss, the model gradually hands over computation to PHM while keeping internal features and logits close to the original. After this residual adaptation, the dense FFNs are removed, and the PHM-only model is briefly fine-tuned. This reduces parameters and FLOPs while preserving the model's captioning behavior.}

    \label{fig:fig2}
\end{figure*}

\section{Method}
\noindent \textbf{Problem Formulation: } Let a pretrained MLLM be parameterized by $\Theta = \{\mathbf{W}_1, \dots, \mathbf{W}_L\}$, where each $\mathbf{W}_\ell \in \mathbb{R}^{d_{\text{out}}^{(\ell)} \times d_{\text{in}}^{(\ell)}}$ is a dense linear transformation with parameter cost $|\mathbf{W}_\ell| = d_{\text{out}}^{(\ell)} \cdot d_{\text{in}}^{(\ell)}$. The total parameter count is therefore $|\Theta| = \sum_{\ell=1}^L d_{\text{out}}^{(\ell)} d_{\text{in}}^{(\ell)}$, which scales quadratically with hidden dimension and dominates both memory and FLOPs. Our objective is to construct a parameter-efficient surrogate $\Theta' = \{\mathbf{W}'_1, \dots, \mathbf{W}'_L\}$ such that each dense matrix $\mathbf{W}_\ell$ is replaced with a structured low-capacity factorization $\mathbf{W}'_\ell = g_{\phi_\ell}(A_\ell, H_\ell)$, where $A_\ell \in \mathbb{R}^{\frac{d_{\text{out}}^{(\ell)}}{2}\times \frac{d_{\text{in}}^{(\ell)}}{2}}$ are learnable factors, $H_\ell$ are fixed basis matrices, and $g_{\phi_\ell}(\cdot)$ denotes our proposed composition rule. The parameter budget is then $|\Theta'| = \sum_{\ell=1}^L |A_\ell| \ll |\Theta|$. Formally, given training data $\mathcal{D} = \{(x_i,y_i)\}_{i=1}^N$, the goal is to solve
$$
\min_{\Theta'} \;\; \frac{1}{N}\sum_{i=1}^N \mathcal{L}(f_{\Theta'}(x_i), y_i) 
\quad \text{subject to} \quad |\Theta'| \leq \rho |\Theta| ,
$$
for a compression ratio $\rho \in (0,1)$, ensuring that the compact model $f_{\Theta'}$ achieves predictive performance comparable to the dense $f_{\Theta}$. Detailed overview of our pipeline is given in Fig.\ref{fig:fig2}.

\subsection{Preliminaries}
\noindent\textbf{Dense Linear Operators and the Parameter Bottleneck: }Consider a multimodal transformer model parameterized by a collection of dense linear mappings $\Theta=\{\mathbf{W}_\ell,\mathbf{b}_\ell\}_{\ell=1}^{L}$. Each feed-forward sublayer applies an affine transformation to hidden states $\mathbf{X}\in \mathbb{R}^{N\times d^{(\ell)}_{\text{in}}}$: $
\mathbf{Y} = \mathbf{X}\mathbf{W}_\ell^\top + \mathbf{b}_\ell,
\qquad \mathbf{W}_\ell \in \mathbb{R}^{d^{(\ell)}_{\text{out}} \times d^{(\ell)}_{\text{in}}}.
$
The parameter cost of this operator is $|\mathbf{W}_\ell| = d^{(\ell)}_{\text{out}} d^{(\ell)}_{\text{in}}$. Summed across layers, the total parameter count becomes $
|\Theta| = \sum_{\ell=1}^L d^{(\ell)}_{\text{out}} d^{(\ell)}_{\text{in}},
$ which scales quadratically with hidden width and dominates both storage and training-time computation. In multimodal large language models, where FFN layers are wide, these matrices constitute the primary bottleneck for parameter reduction and FLOP efficiency.

\noindent \textbf{PHM Reparameterization of Large Dense Linears: }To address the quadratic growth of dense linear operators in multimodal transformers, we substitute large matrices by a PHM structure. This representation factorizes a $2n \times 2m$ mapping into Kronecker products of small $2\times2$ hypercomplex bases with learnable $n \times m$ matrices, thereby reducing parameter count while maintaining expressive power.\\
\textit{(1) Residual PHM Substitution: }For each eligible layer $\ell$, whose size exceeds a threshold $\tau$, the dense operator $\mathbf{W}_\ell$ is reparameterized by a residual formulation:

$$
\mathbf{W}^{\mathrm{res}}_\ell(\alpha_\ell)
= (1-\alpha_\ell)\,\mathbf{W}_\ell
+ \alpha_\ell \underbrace{\sum_{b=1}^{B_\ell}\!\big(H_b^{(\ell)}\otimes A_b^{(\ell)}\big)}_{\mathbf{W}_\ell^{\mathrm{PHM}}}
$$ where $H_b^{(\ell)} \in \mathbb{R}^{2\times2}$ are fixed hypercomplex bases, $A_b^{(\ell)} \in \mathbb{R}^{\tfrac{d^{(\ell)}_{\text{out}}}{2}\times \tfrac{d^{(\ell)}_{\text{in}}}{2}}$ are learnable matrices, and $\alpha_\ell \in [0,1]$ interpolates between the dense operator and its PHM substitute. This formulation ensures a smooth transition during training, enabling the PHM representation to inherit knowledge from the dense operator.\\
\textit{(2) Parameter Capacity: } The substitution yields substantial parameter reduction. A dense operator requires $
|\mathbf{W}_\ell| = d^{(\ell)}_{\text{out}}d^{(\ell)}_{\text{in}},
$ whereas the PHM operator requires $
|\mathbf{W}_\ell^{\mathrm{PHM}}| = B_\ell\,\frac{d^{(\ell)}_{\text{out}}d^{(\ell)}_{\text{in}}}{4}.
$ Hence, the compression factor is approximately $4/B_\ell$. For $B_\ell=2$, this corresponds to a two-fold reduction, while $B_\ell=3$ provides a reduction by a factor of $1.33$. This formulation allows precise control over the representational capacity allocated to each layer.\\
\textit{(3) Block Form and Canonical Bases: } Let the input and output dimensions satisfy $d_{\text{in}}=2m,\ d_{\text{out}}=2n$. Each Kronecker term expands as
$$
H_b \otimes A_b =
\begin{bmatrix}
h_{11}^{(b)}A_b & h_{12}^{(b)}A_b \\
h_{21}^{(b)}A_b & h_{22}^{(b)}A_b
\end{bmatrix}.
$$
Two canonical cases clarify the construction. In the case $B=2$, using the canonical bases $H_1 = I = \begin{bmatrix}1&0\\0&1\end{bmatrix}$ and $H_2 = J = \begin{bmatrix}0&-1\\1&0\end{bmatrix}$, the PHM operator is $
\mathbf{W}^{\text{PHM}} = 
\begin{bmatrix}
A_1 & -A_2 \\
A_2 & A_1
\end{bmatrix},
$ which corresponds to a complex-like factorization. Extending to $B=3$ by including the basis $H_3 = \begin{bmatrix}1&0\\0&-1\end{bmatrix}$, the operator becomes
$$
\mathbf{W}^{\text{PHM}} = 
\begin{bmatrix}
A_1 + A_3 & -A_2 \\
A_2 & A_1 - A_3
\end{bmatrix},
$$
thereby enlarging the representational subspace and capturing anisotropy not represented by the $B=2$ case.

\noindent\textbf{Dense-to-PHM Initialization: }To ensure stable training, PHM parameters are initialized by projecting pretrained dense weights into the PHM subspace. Partition the dense weight $\mathbf{W}$ into four blocks $W_{(u,v)} \in \mathbb{R}^{n\times m}$, with $u,v\in\{1,2\}$. The projection problem is $
\min_{\{A_b\}} \Big\| \mathbf{W} - \sum_{b=1}^{B} H_b \otimes A_b \Big\|_F^2.
$ Vectorization yields a compact linear system
$$
\scriptsize
\underbrace{\begin{bmatrix}
h_{11}^{(1)} & \cdots & h_{11}^{(B)} \\
h_{12}^{(1)} & \cdots & h_{12}^{(B)} \\
h_{21}^{(1)} & \cdots & h_{21}^{(B)} \\
h_{22}^{(1)} & \cdots & h_{22}^{(B)}
\end{bmatrix}}_{\mathsf{H}\in\mathbb{R}^{4\times B}}
\underbrace{\begin{bmatrix}\mathrm{vec}(A_1)\\ \vdots \\ \mathrm{vec}(A_B)\end{bmatrix}}_{a}
\approx
\underbrace{\begin{bmatrix}
\mathrm{vec}(W_{(1,1)}) \\ \mathrm{vec}(W_{(1,2)}) \\ \mathrm{vec}(W_{(2,1)}) \\ \mathrm{vec}(W_{(2,2)})
\end{bmatrix}}_{w},
\quad a \approx (\mathsf{H}^\dagger \otimes I)\, w.
$$ For $B=2$ with canonical $(I,J)$ bases, if the block decomposition yields submatrices $p,q,r,s$, the closed-form initialization is $
A_1 = \tfrac{1}{2}(p+s), \qquad A_2 = \tfrac{1}{2}(r-q).
$ For $B\geq 3$, the least-squares solution is employed. This guarantees that the PHM approximation $\mathbf{W}_\ell^{\mathrm{PHM}}$ remains close to the dense $\mathbf{W}_\ell$ in Frobenius norm.

\noindent \textbf{Selective Capacity Assignment: }Let the language stack contain $L_{\text{lang}}$ transformer layers indexed $\ell=0,\dots,L_{\text{lang}}-1$. Denote by $\mathcal{S}$ the set of FFN linears selected for PHM substitution. A layer-specific basis count $B_\ell\in\{2,3\}$ is assigned according to proximity to the output head: with a threshold $t^\star=L_{\text{lang}}-K$ (for a fixed $K\ge 0$),
$$
B_\ell \;=\;
\begin{cases}
3, & \text{if } \ell\in\mathcal{S}\cap\{\;t^\star,\,t^\star\!+\!1,\,\dots,\,L_{\text{lang}}\!-\!1\;\},\\[4pt]
2, & \text{if } \ell\in\mathcal{S}\setminus\{\;t^\star,\,\dots,\,L_{\text{lang}}\!-\!1\;\}.
\end{cases}
$$ Attention projections (Q/K/V and output-proj) remain dense, and the multimodal projector is retained dense unless explicitly swapped. The assignment follows a sensitivity rationale: linearization around current parameters yields the output Jacobian $J_\ell=\partial \mathrm{logits}/\partial \mathbf{W}_\ell$ and the associated empirical Fisher $F_\ell\simeq \mathbb{E}[J_\ell^\top J_\ell]$, which concentrates toward top-of-stack language layers. Under a PHM factorization with $B$ bases, the admissible subspace $\mathcal{U}_B=\{\sum_{b=1}^B H_b\otimes A_b\}\subset\mathbb{R}^{d^{(\ell)}_{\text{out}}\times d^{(\ell)}_{\text{in}}}$ has dimension $B\,\tfrac{d^{(\ell)}_{\text{out}}d^{(\ell)}_{\text{in}}}{4}$; the best Frobenius approximation error of a dense $\mathbf{W}_\ell$ by $\mathcal{U}_B$ decreases monotonically with $B$. A layerwise risk proxy under small perturbations is $\mathbb{E}[\Delta \mathcal{L}_\ell]\approx \tfrac{1}{2}\langle \Delta \mathbf{W}_\ell,\,F_\ell\,\Delta \mathbf{W}_\ell\rangle$; allocating larger $B_\ell$ where $\mathrm{tr}(F_\ell)$ is larger reduces this proxy for a fixed parameter budget. The allocation can be expressed as a discrete budgeted program:
$$
\min_{\{B_\ell\in\{2,3\}\}}\ \sum_{\ell\in \mathcal{S}} \Phi_\ell\big(B_\ell\big)
\quad\text{s.t.}\quad 
\sum_{\ell\in \mathcal{S}} B_\ell n_\ell m_\ell \;\le\; \mathcal{B},
$$ where $d^{(\ell)}_{\text{in}}=2m_\ell$, $d^{(\ell)}_{\text{out}}=2n_\ell$, $\mathcal{B}$ is a global parameter budget, and $\Phi_\ell(B)$ is a decreasing surrogate of approximation risk, e.g. $\Phi_\ell(B)\propto \mathrm{tr}(F_\ell)\,\epsilon_\ell(B)$ with $\epsilon_\ell(B)$ the best-subspace error induced by $\mathcal{U}_B$. The heuristic $B_\ell=3$ on the top $K$ language MLP layers and $B_\ell=2$ elsewhere implements a monotone allocation aligned with $\mathrm{tr}(F_\ell)$ and achieves a favorable loss-per-parameter trade-off. This yields higher capacity where logits are most sensitive while preserving near $2\times$ compression in the remaining stack.

\subsection{Residual Transition }The direct replacement of dense operators with PHM substitutes may lead to instability during optimization. To mitigate this, training employs a residual transition governed by the interpolation coefficient $\alpha_\ell$, which is scheduled to increase gradually over training steps. Formally, one sets $\alpha_\ell(t) = \min(1, t/T_{\text{fade}})$, where $T_{\text{fade}}$ is a predetermined fade horizon. This ensures that, at the start of training, the network behaves identically to the dense teacher, and gradually evolves into a PHM-based student. The optimization process is driven by three complementary objectives. The first is the label-smoothed cross-entropy, which accounts for supervised prediction accuracy:
$$
\mathcal{L}_{\text{CE}}=-\frac{1}{Z}\sum_{i,t}\Big[(1-\varepsilon)\log p_{\Theta(\alpha)}(y_{i,t})+$$
$$\frac{\varepsilon}{V-1}\sum_{v\ne y_{i,t}}\log p_{\Theta(\alpha)}(v)\Big],
$$
where $\varepsilon$ is the smoothing parameter and $V$ the vocabulary size. The second is a knowledge distillation term, defined with respect to the dense teacher network ($\alpha\equiv 0$), which transfers predictive behavior from the teacher to the PHM student:
$$
\mathcal{L}_{\text{KD}}=\frac{T^2}{Z}\sum_{i,t}\mathrm{KL}\!\left(\mathrm{softmax}\!\tfrac{z^T_{i,t}}{T}\;\middle\|\;\mathrm{softmax}\!\tfrac{z^S_{i,t}}{T}\right),
$$
where $z^T$ and $z^S$ denote the teacher and student logits, respectively, and $T$ is a temperature parameter. The third is a reconstruction penalty, which ensures that PHM operators remain close to their dense counterparts during transition:
$$
\mathcal{L}_{\text{recon}}=\frac{1}{|\mathcal{M}|}\sum_{\ell\in\mathcal{M}}\mathbb{E}_x\big\|\mathbf{W}_\ell^{\text{PHM}}x-\mathbf{W}_\ell^{\text{dense}}x\big\|_2^2.
$$
The complete objective function is then expressed as
$
\mathcal{L}(t)=(1-\lambda(t))\mathcal{L}_{\text{CE}}+\lambda(t)\mathcal{L}_{\text{KD}}+\mu\,\mathcal{L}_{\text{recon}},
$ where $\lambda(t)=\lambda_{\max}\min(1,t/T_{\text{fade}})$ and $\mu\ll1$. This formulation ensures that the transition from dense to PHM is smooth, that task performance is preserved via CE, that function-level alignment is maintained through KD, and that operator-level approximation is penalized through the reconstruction term.

\subsection{Adaptation and Complexity Analysis}
The training pipeline is structured in two stages to balance stability and efficiency. In \textit{Stage A (Residual Adaptation)}, training starts with $\alpha=0,\lambda=0$, so the model behaves like the dense teacher and is optimized with cross-entropy. As steps progress, $\alpha\!\to\!1$ and $\lambda\!\to\!\lambda_{\max}$, gradually shifting capacity to PHM while transferring predictive behavior from the dense teacher. At the end, all operators are PHM ($\alpha=1$), and the best snapshot is selected on validation:
\(
\Theta^\star=\arg\max_{\text{snapshots}} g(f_{\Theta'};\mathcal{D}_{\text{val}}), \qquad \Theta'=\text{PHM-only form}.
\)
In \textit{Stage B (PHM-only Fine-Tuning)}, dense paths are removed and optimization continues only over PHM and the head:
\(
\min_{\Theta'} \mathcal{L}_{\text{CE}}(\Theta') \quad \text{s.t.}\quad |\Theta'|\le \rho|\Theta|, \ \rho\in(0,1).
\)

The complexity benefits are explicit. For $d_{\text{in}}=2m,\ d_{\text{out}}=2n$, dense FFNs require $4nm$ parameters, while PHM requires $Bnm$. With selective capacity ($B_\ell=3$ for top $K$ language MLPs, $B_\ell=2$ elsewhere), the total becomes
\(
|\Theta'|=\sum_{\ell\notin\mathcal{S}}4n_\ell m_\ell+\sum_{\ell\in\mathcal{S}}B_\ell n_\ell m_\ell+\sum_{\text{text-attn LoRA}}2r(d_{\text{in}}+d_{\text{out}})+|\text{proj+head}|,
\)
where $\mathcal{S}$ are swapped layers. Computationally, dense operators cost $\approx 2N d_{\text{in}}d_{\text{out}}$, while PHM reduces this to
\[
\text{FLOPs}_{\text{PHM}} \approx BNd_{\text{in}}d_{\text{out}}+O(N(d_{\text{in}}+d_{\text{out}})),
\]
showing substantial parameter savings with FLOPs no worse (and often lower) for $B_l<4$. Stage A thus ensures a smooth, knowledge-preserving transition, while Stage B refines the compact PHM model directly on the task objective.

\begin{algorithm}[!t]
\scriptsize
\caption{Residual PHM Compression with Selective Capacity, KD, and LoRA}
\label{algo}
\begin{algorithmic}[1]
\Require Pretrained MLLM $\Theta=\{\mathbf{W}_\ell,\mathbf{b}_\ell\}_{\ell=1}^{L}$; language depth $L_{\mathrm{lang}}$; top budget $K$; size threshold $\tau$; bases $\{H_b\}_{b=1}^{B_{\max}}$ (canonical $I,J,(K)$); fade $T_{\mathrm{fade}}$; KD ramp $\lambda(t)$ with $\lambda_{\max}$; temperature $T$; smoothing $\varepsilon$; recon weight $\mu\ll1$; LoRA rank $r$ and scale $\gamma$.
\Ensure PHM-only $\Theta'$ after Stage~B.
\vspace{2pt}

\Statex \textbf{Phase 0: Structural preparation (single pass)}
\State $\mathcal{S}\!\gets\!\{\ell:\ d_{\mathrm{out}}^{(\ell)}d_{\mathrm{in}}^{(\ell)}\!\ge\!\tau,\ \ell\ \text{is FFN, not attn/emb/head}\}$
\State $t^\star\!\gets\!L_{\mathrm{lang}}-K$
\For{$\ell\in\mathcal{S}$}
  \State $B_\ell\!\gets\!\begin{cases}3, & \ell\ \text{is language MLP and }\ell\ge t^\star\\ 2,&\text{otherwise}\end{cases}$
  \State Factor $d_{\mathrm{in}}^{(\ell)}\!=\!2m,\ d_{\mathrm{out}}^{(\ell)}\!=\!2n$; create $A_b^{(\ell)}\!\in\!\mathbb{R}^{n\times m}$
  \State Residual: $\mathbf{W}^{\mathrm{res}}_\ell(\alpha)=(1-\alpha)\mathbf{W}_\ell+\alpha\sum_{b=1}^{B_\ell}(H_b\!\otimes\!A_b^{(\ell)})$
  \State Dense$\to$PHM init: partition $\mathbf{W}_\ell\!\to\!\{p,q,r,s\}$
  \State \textbf{if} $B_\ell\!=\!2$ (bases $I,J$) \textbf{then} $A_1^{(\ell)}\!\gets\!(p+s)/2,\ A_2^{(\ell)}\!\gets\!(r-q)/2$ \textbf{else} LSQ on blocks
\EndFor
\For{language self-attn $q,k,v$}
  \State LoRA: $\Delta\mathbf{W}\!=\!\gamma\,\mathbf{B}\mathbf{A}$; freeze base; $\mathbf{W}'\!\gets\!\mathbf{W}+\Delta\mathbf{W}$
\EndFor

\vspace{2pt}
\Statex \textbf{Phase 1: Stage A (residual adaptation + KD + recon)}
\State $t\!\gets\!0$;\ \ $\alpha_\ell(0)\!\gets\!0$ for $\ell\!\in\!\mathcal{S}$;\ \ supervised token set $\mathcal{T}\!=\!\{(i,u):y_{i,u}\!\neq\!-100\}$
\While{training}
  \State $\alpha_\ell(t)\!=\!\min(1,\ t/T_{\mathrm{fade}})$,\ \ $\lambda(t)\!=\!\lambda_{\max}\,\alpha_\ell(t)$
  \State \textbf{Student:} forward with $\mathbf{W}^{\mathrm{res}}_\ell(\alpha_\ell(t))$ $\Rightarrow$ logits $z^S$
  \State \textbf{Teacher:} temporarily set $\alpha_\ell\!\equiv\!0$ on residual blocks; same batch $\Rightarrow$ logits $z^T$
  \State Loss on $\mathcal{T}$:
  \Statex \hspace{0.6em} CE: $\mathcal{L}_{\mathrm{CE}}=\mathrm{CE}_\varepsilon(z^S,y)$;\quad KD: $\mathcal{L}_{\mathrm{KD}}=\mathrm{KD}_T(z^S,z^T)$;\quad Recon: $\mathcal{L}_{\mathrm{recon}}=\frac{1}{|\mathcal{M}|}\!\sum_{\ell\in\mathcal{M}}\!\mathbb{E}\|W_\ell^{\mathrm{PHM}}x-W_\ell^{\mathrm{dense}}x\|_2^2$
  \State Objective: $\mathcal{L}(t)=(1-\lambda(t))\mathcal{L}_{\mathrm{CE}}+\lambda(t)\mathcal{L}_{\mathrm{KD}}+\mu\,\mathcal{L}_{\mathrm{recon}}$
  \State Update $\{A_b^{(\ell)}\}$, LoRA, projector (if unfrozen), LM head; dense FFNs frozen
  \State $t\!\gets\!t+1$
\EndWhile
\State Collapse: set $\alpha_\ell\!=\!1$; drop dense branches $\Rightarrow$ PHM-only $f_{\Theta'}$

\vspace{2pt}
\Statex \textbf{Phase 2: Selection (PHM-only validation)}
\State Pick snapshot by metric $g$ (e.g., lowest CE$_{\mathrm{PHM}}$ or highest CIDEr): $\ \Theta^\star\in\arg\max g\big(f_{\Theta'};\mathcal{D}_{\mathrm{val}}\big)$

\vspace{2pt}
\Statex \textbf{Phase 3: Stage B (PHM-only fine-tuning)}
\State Initialize with $\Theta^\star$ (no residual path); optimize $\min_{\Theta'}\ \mathrm{CE}_\varepsilon$ on $\mathcal{T}$ (optional light KD); merge LoRA for deploy

\vspace{2pt}
\Statex \textbf{Accounting (per swapped $\ell\in\mathcal{S}$)}
\State Params: dense $=4nm$, PHM $=B_\ell nm$ (compression $\approx 4/B_\ell$);\quad FLOPs (per batch of $N$ vecs): dense $\approx 2N d_{\mathrm{in}}d_{\mathrm{out}}$, PHM $\approx B_\ell N d_{\mathrm{in}}d_{\mathrm{out}}$
\end{algorithmic}
\end{algorithm}

\section{Experiment}
\subsection{Setup}
\noindent \textbf{Dataset:} We use five multimodal datasets spanning two tasks. For \textit{caption generation}, we use Flickr30k~\cite{plummer2015flickr30k}, NoCaps~\cite{agrawal2019nocaps}, and COCO-Cap~\cite{lin2014microsoft}. Flickr30k and COCO-Cap serve as training sources, while NoCaps is reserved exclusively for assessing generalization to novel object categories. For \textit{visual question answering}, we evaluate on ScienceQA~\cite{lu2022learn} and FinMME~\cite{luo2025finmme}, which require multimodal reasoning in scientific and financial domains, respectively. This configuration allows the model to be trained solely on captioning corpora while being evaluated on both captioning and QA tasks under a unified framework. All datasets are converted into a consistent instruction-following conversational format. Each sample $(I_i,q_i,y_i)$, comprising an image $I_i$, a user query $q_i$, and a ground-truth target $y_i$, is mapped to a structured dialogue, \(
\mathcal{C}_i = \big[(\texttt{user},\{I_i, q_i\}), \; (\texttt{assistant}, y_i)\big],
\) which is then serialized into a multimodal token sequence $x_i=(x_{i,1},\dots,x_{i,L_i})$. A supervision mask $\ell_{i,t}$ ensures that only assistant tokens contribute to the training objective, with $\ell_{i,t}=-100$ for image and user tokens, and $\ell_{i,t}=x_{i,t}$ for assistant tokens. 

\noindent A Flickr30k captioning instance is represented as:
\begin{tcolorbox}[title=Chat-formatted Example, colback=gray!5, colframe=black]
\scriptsize
\begin{verbatim}
<im_start>user
Picture: <img> /image.jpg </img> 
Describe the picture <im_end>
<im_start>assistant
The sign is a road closure with an orange 
rhombus. <im_end>
\end{verbatim}
\end{tcolorbox}

\begin{table*}[t]
  \caption{Performance comparison on image captioning benchmarks across Flickr, NoCaps, and COCO-Cap. B@1 and B@4 denote BLEU-1 and BLEU-4, M denotes METEOR, and R denotes ROUGE. The best scores for each column are highlighted in \textcolor{red}{\textbf{red bold}}, and the second-best scores in \textcolor{blue}{\textbf{blue bold}}. Our PHM-compressed variants reduce the parameter counts of 7B-scale models while maintaining competitive performance: InstructBLIP-7B ($\sim$5.45B vs.\ 7B), LLaVA-1.5-7B ($\sim$5.15B vs.\ 7B), and Qwen2.5-VL-7B ($\sim$5.38B vs.\ 7B).}
 \label{tab:caption_results}
 \centering
\small
\setlength{\tabcolsep}{3pt}
\begin{adjustbox}{max width=\textwidth}
\begin{tabular}{@{}l *{15}{c}@{}}
    \toprule
    \multirow{2}{*}{Model} & \multicolumn{5}{c}{Flickr (31,784)} & \multicolumn{5}{c}{NoCaps (4,500)} & \multicolumn{5}{c}{COCO-Cap (5,000)} \\
    \cmidrule(lr){2-6} \cmidrule(lr){7-11} \cmidrule(l){12-16}
    & CIDEr & B@1 & B@4 & M & R & CIDEr & B@1 & B@4 & M & R & CIDEr & B@1 & B@4 & M & R \\
    \midrule
    InstructBLIP-7B~\cite{dai2023instructblip} & 78.2 & \textcolor{blue}{\textbf{77.1}} & \textcolor{blue}{\textbf{30.9}} & 24.8 & 53.4 & 118.2 & 88.2 & \textcolor{red}{\textbf{47.2}} & 30.3 & \textcolor{blue}{\textbf{62.0}} & \textcolor{red}{\textbf{141.3}}  & \textcolor{red}{\textbf{82.8}}  & \textcolor{red}{\textbf{41.7}}  & \textcolor{red}{\textbf{30.9}}  & \textcolor{red}{\textbf{61.0}}  \\
    InstructBLIP-13B~\cite{dai2023instructblip} & 76.1 & 76.8 & 30.1 & 24.4 & 53.0 & 116.3 & \textcolor{blue}{\textbf{88.3}} & 46.7 & 29.8 & 61.4 & 135.0 & \textcolor{blue}{\textbf{82.2}}  & 39.7  & 29.8  & 59.8  \\
    LLaVA-1.5-7B~\cite{liu2024improved} & 74.9 & 71.7 & 28.4 & 26.1 & 52.8 & 105.5 & 82.6 & 40.2 & 30.3 & 59.4 & 110.3 & 73.0 & 29.7 & 29.2 & 55.5 \\
    LLaVA-1.5-13B~\cite{liu2024improved} & 79.4 & 73.6 & 30.2 & \textcolor{blue}{\textbf{26.6}} & \textcolor{red}{\textbf{53.9}} & 109.2 & 84.2 & 42.4 & \textcolor{red}{\textbf{30.6}} & 60.3 & 115.6 & 74.6 & 31.5 & 29.4 & 56.5 \\
    LLaVA-1.6-7B~\cite{liu2024llava} & 68.4 & 69.6 & 26.6 & 23.2 & 50.3 & 88.4 & 73.8 & 34.8 & 25.9 & 54.6 & 99.9 & 67.7 & 28.4 & 25.5 & 52.4 \\
    LLaVA-1.6-13B~\cite{liu2024llava} & 66.6 & 65.2 & 24.2 & 22.2 & 48.8 & 88.1 & 68.7 & 34.0 & 25.4 & 54.9 & 101.8 & 62.2 & 27.5 & 24.6 & 52.1 \\
    MiniCPM-V-3B~\cite{hu2023large} & 66.8 & 68.0 & 25.1 & \textcolor{red}{\textbf{27.2}} & 51.0 & 89.9 & 79.1 & 33.2 & 29.7 & 55.8 & 94.2 & 69.8 & 23.9 & 28.3 & 52.3 \\
    DeCap~\cite{hu2023large} & 56.7 &  -- & 21.2 & 21.8 & --  & 42.7 & --  &  -- & --  & --  & 91.2 & --  & 24.7 & 25.0 & --  \\
    Qwen2.5 VL-7B~\cite{bai2025qwen2} & \textcolor{red}{\textbf{85.8}} & --  &  -- & --  & --  & \textcolor{blue}{\textbf{121.4}} & --  &  -- & --  & --  & 89.4 & --  & --  & --  & --  \\
    \midrule
    InstructBLIP-7B + PHM (Ours) & 77.0 & 76.0 & 30.2 & 24.5 & 53.0 & 116.8 & 87.6 & 46.5 & 29.9 & 61.5 & \textcolor{blue}{\textbf{138.2}} & 82.0 & \textcolor{blue}{\textbf{40.9}} & \textcolor{blue}{\textbf{30.3}} & \textcolor{blue}{\textbf{60.4}} \\
    LLaVA-1.5-7B + PHM (Ours) & 73.5 & 70.8 & 27.9 & 25.7 & 52.3 & 104.0 & 81.8 & 39.6 & 29.8 & 58.7 & 108.9 & 72.2 & 29.1 & 28.8 & 55.0 \\
    Qwen2.5 VL-7B + PHM (Ours) & \textcolor{blue}{\textbf{85.2}} & \textcolor{red}{\textbf{78.8}} & \textcolor{red}{\textbf{31.0}} & 25.0 & \textcolor{blue}{\textbf{53.6}} & \textcolor{red}{\textbf{122.1}} & \textcolor{red}{\textbf{89.2}} & \textcolor{blue}{\textbf{47.1}} & \textcolor{blue}{\textbf{30.4}} & \textcolor{red}{\textbf{62.3}} & 89.0 & 73.5 & 30.1 & 29.6 & 55.7 \\
    \bottomrule
\end{tabular}
\end{adjustbox}
\end{table*}

\begin{table}[!t]
\centering
\caption{Evaluation on ScienceQA and FinMME benchmarks. The best scores in each column are highlighted in \textcolor{red}{\textbf{red bold}}, and the second-best in \textcolor{blue}{\textbf{blue bold}}. Our PHM-compressed models reduce 7B-scale parameters to 5.15–5.45B (LLaVA, InstructBLIP, Qwen) while maintaining competitive accuracy, with only modest drops and strong performance, especially on SciQA.}
\setlength{\tabcolsep}{4pt}
\begin{adjustbox}{max width=\columnwidth}
\begin{tabular}{@{}lcc@{}}
    \toprule
    Model & SciQA-IMG~\cite{bai2025qwen2} & FinMME~\cite{luo2025finmme} \\
    \midrule
    BLIP-2 (Vicuna-7B)~\cite{dai2023instructblip} & 53.81 & --  \\
    BLIP-2 (Vicuna-13B)~\cite{dai2023instructblip} & 61.01 &  -- \\
    InstructBLIP (FianT5xL)~\cite{dai2023instructblip} & \textcolor{red}{\textbf{70.40}} &  -- \\
    InstructBLIP (Vicuna-7B)~\cite{dai2023instructblip} & 60.52 & 24.50  \\
    InstructBLIP (Vicuna-13B)~\cite{dai2023instructblip} & 63.12 &  26.21 \\
    DeepSeekVL-2 Small~\cite{wu2024deepseek} &  -- & 34.14 \\
    Qwen2.5-VL 72B~\cite{bai2025qwen2} &  \textcolor{blue}{\textbf{70.31}} & \textcolor{red}{\textbf{49.64}} \\
    Qwen2-VL 72B~\cite{bai2025qwen2} &  68.12 & \textcolor{blue}{\textbf{37.11}} \\
    Qwen2-VL 7B~\cite{bai2025qwen2} & 63.51  & 34.14 \\
    Qwen2.5-VL 3B~\cite{bai2025qwen2} & 59.22  & 32.53 \\
    \midrule
    InstructBLIP + PHM (Ours) & 46.5 & 21.5 \\
    LLaVA-1.5-7B + PHM (Ours) & 55.2 & 28.5 \\
    Qwen2.5 VL-7B + PHM (Ours) & 60.5 & 33.1 \\
    \bottomrule
\end{tabular}
\end{adjustbox}
\label{tab:vqa_results_simplified}
\end{table}

This unified preprocessing setup ensures that captioning and QA datasets are aligned under a single instruction-following paradigm, enabling joint optimization and consistent evaluation across tasks.

\noindent\textbf{Model Overview:} We evaluate three representative MLLMs that span diverse multimodal architectures. 
\textit{Qwen2.5-VL-7B-Instruct}~\cite{bai2025qwen2} combines a ViT-based vision tower with a Qwen LLM and is instruction-tuned for multimodal reasoning. The Qwen2.5-VL family includes 3B, 7B, and 72B variants, offering a scalable range of capabilities.
\textit{InstructBLIP}~\cite{dai2023instructblip} builds upon BLIP-2 and introduces instruction tuning across 26 multimodal datasets, leveraging a Query Transformer to strengthen cross-modal alignment.
\textit{LLaVA-1.5-7B}~\cite{liu2024improved} combines a CLIP ViT-L image encoder with an MLP projection layer feeding into a Vicuna/LLaMA backbone. Its 1.5 release incorporates improved visual instruct tuning to achieve stronger baselines.

\noindent \textbf{Implementation Details: }For each model $M$, large FFN layers above a threshold $\tau$ are reparameterized into residual PHM, with $B_\ell=2$ generally and $B_\ell=3$ in the top $K$ language layers; initialization uses least-squares projection from dense weights. Attention projections remain dense, with LoRA applied to text-side Q/K/V ($r$, $\alpha_{\text{LoRA}}$). Training proceeds in two stages: Stage~A uses a fade $\alpha_\ell(t)=\min(1,t/T_{\text{fade}})$ and KD ramp $\lambda(t)=\lambda_{\max}\min(1,t/T_{\text{fade}})$, optimizing a loss of label-smoothed cross-entropy ($\varepsilon=0.1$), KD ($T=4$), and a reconstruction penalty. Stage~B removes dense paths and fine-tunes PHM-only layers with CE. Optimization uses AdamW with cosine decay, warmup, weight decay, gradient clipping, and accumulation. Implementation relies on PyTorch, HuggingFace \texttt{transformers}/\texttt{accelerate}, with bfloat16, gradient checkpointing, and FlashAttention2/SDP attention. Data and chat formatting follow HuggingFace multimodal APIs. FLOPs are measured via forward hooks. All experiments are conducted on NVIDIA H100 GPUs. Algorithm~\ref{algo} details the full pipeline.


\noindent \textbf{Evaluation Metrics: }For captioning, we report BLEU-1/4 for $n$-gram precision, ROUGE-L for recall via longest common subsequence, and CIDEr for consensus with human references using TF–IDF weighting. For question answering, accuracy is measured as the proportion of exact matches with ground truth. These metrics jointly capture precision, recall, semantic relevance, and correctness.

\subsection{Baselines}
We evaluate our PHM-compressed multimodal LLMs on three widely used image captioning benchmarks, Flickr, NoCaps, and COCO-Cap, as summarized in Table~\ref{tab:caption_results}. In contrast to prior work that often reports single-dataset outcomes, our analysis spans diverse benchmarks and multiple evaluation metrics (CIDEr, BLEU, METEOR, ROUGE), thereby providing a more holistic view of generalization performance. Baseline results for InstructBLIP and LLaVA closely match those reported in their original publications, confirming the reliability of our reproduction pipeline. Among open-source baselines, InstructBLIP-7B obtained strong results on COCO-Cap (CIDEr 141.3), while LLaVA-1.5 consistently outperformed LLaVA-1.6 across datasets. By contrast, MiniCPM-V and DeCap underperformed on generalization, reflecting weaker alignment with captioning objectives. Proprietary models such as Flamingo, GPT-4V, and Gemini reported partial results, but did not consistently outperform the strongest open-source baselines.  

\subsection{Results}
Our PHM-compressed models demonstrate competitive performance relative to their uncompressed counterparts. Qwen2.5-VL-7B (PHM) maintains strong results across all datasets, achieving near-parity with the base model on Flickr30k (CIDEr 85.2 vs.\ 85.8) and COCO-Cap (89.0 vs.\ 89.4), and even slightly surpassing it on NoCaps (122.1 vs.\ 121.4). InstructBLIP-7B (PHM) achieves CIDEr 138.2 on COCO-Cap compared to 141.3 for the dense model, while LLaVA-1.5-7B (PHM) obtains 108.9 versus 110.3. Overall, PHM substitution yields only modest reductions in accuracy while preserving strong competitiveness. To evaluate generalization beyond captioning, we assess ScienceQA and FinMME (Table~\ref{tab:vqa_results_simplified}). InstructBLIP (FiT5xL) achieves the strongest ScienceQA performance (70.4), while Qwen2.5-VL-72B leads FinMME accuracy (49.64). Our PHM-compressed Qwen2.5-VL-7B achieves 60.5 on ScienceQA and 33.1 on FinMME, trailing larger baselines but demonstrating robust cross-task performance despite reduced capacity. Finally, we quantify the efficiency gains introduced by PHM. By reparameterizing large dense operators, PHM reduces the parameter counts of 7B-class models substantially: from 7B to 5.38B for Qwen2.5, from 7B to 5.15B for LLaVA-1.5, and from 7B to 5.45B for InstructBLIP. These reductions correspond to 20–30\% savings in parameters and FLOPs, while preserving downstream performance.

\noindent \textbf{Observations:} The findings above highlight three key takeaways: 
(a) PHM enables substantial parameter reduction while retaining strong captioning and QA performance, demonstrating its viability as a compression strategy for multimodal LLMs. (b) Selective allocation of higher PHM capacity ($B_l=3$) in top language layers contributes to stabilizing accuracy, particularly on sensitive benchmarks such as NoCaps. (c) The trade-off between efficiency and performance remains favorable: PHM-compressed models operate at reduced FLOPs and inference latency, while maintaining accuracy within $1$–$3\%$ of base models.

\subsection{Ablation Study}

\begin{figure*}[t]
    \centering
    \begin{subfigure}[b]{0.25\textwidth}
        \centering
        \includegraphics[width=\linewidth, trim=10 10 10 10, clip]{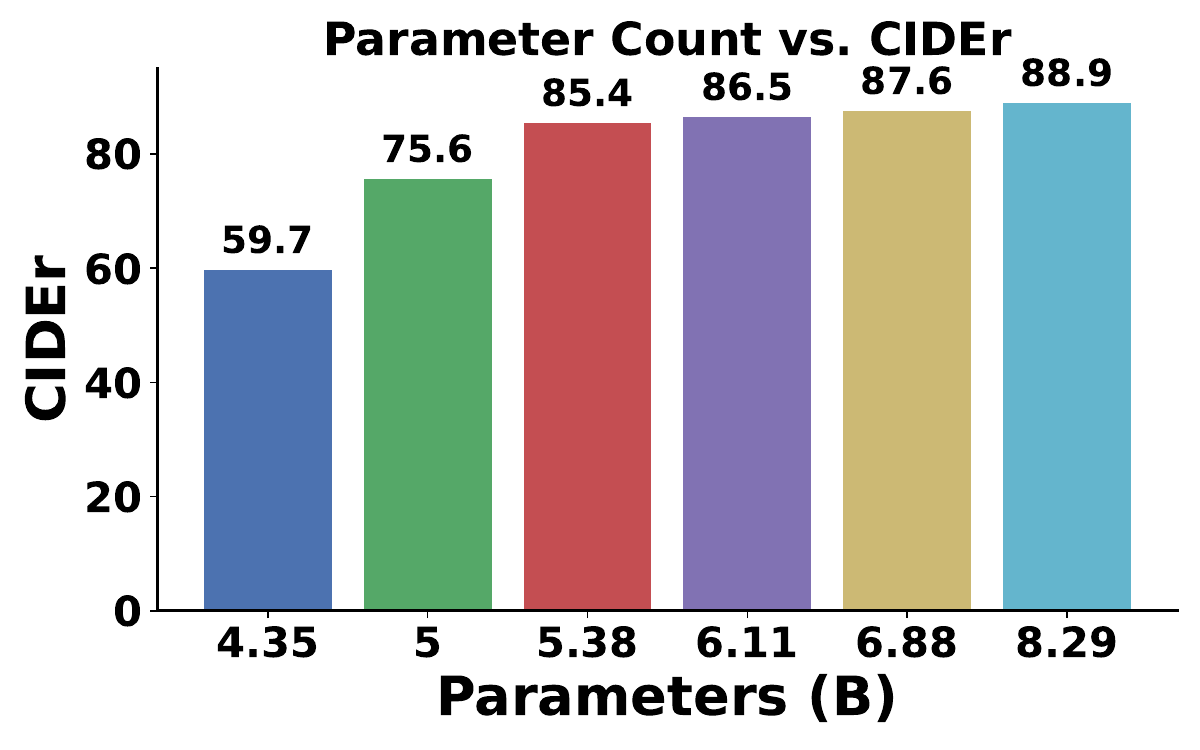}
        \caption{Parameter count vs. CIDEr.}
        \label{fig:param_cider}
    \end{subfigure}
    \hfill
    \begin{subfigure}[b]{0.25\textwidth}
        \centering
        \includegraphics[width=\linewidth, trim=10 10 10 10, clip]{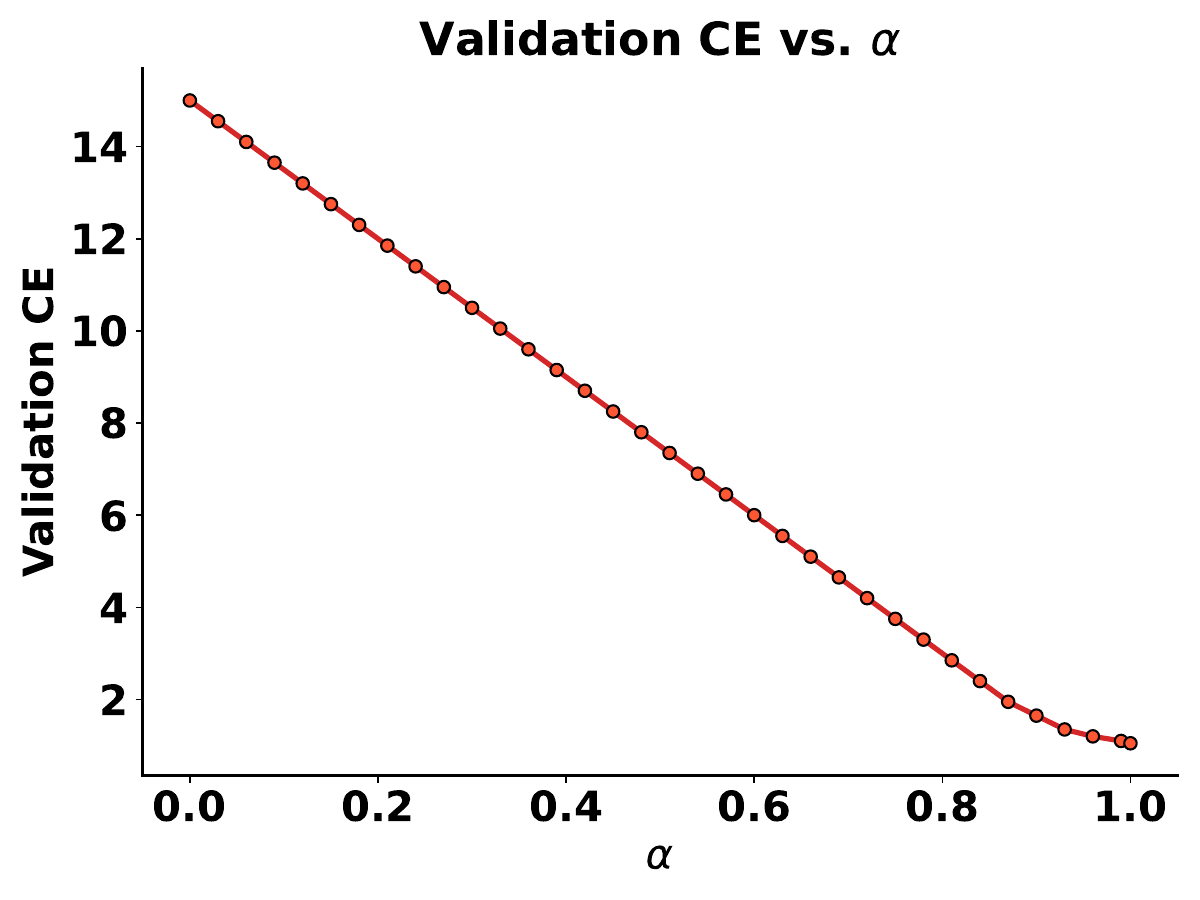}
        \caption{$\alpha$ vs. validation CE.}
        \label{fig:alpha_valce}
    \end{subfigure}
    \hfill
    \begin{subfigure}[b]{0.23\textwidth}
        \centering
        \includegraphics[width=\linewidth, trim=10 10 10 10, clip]{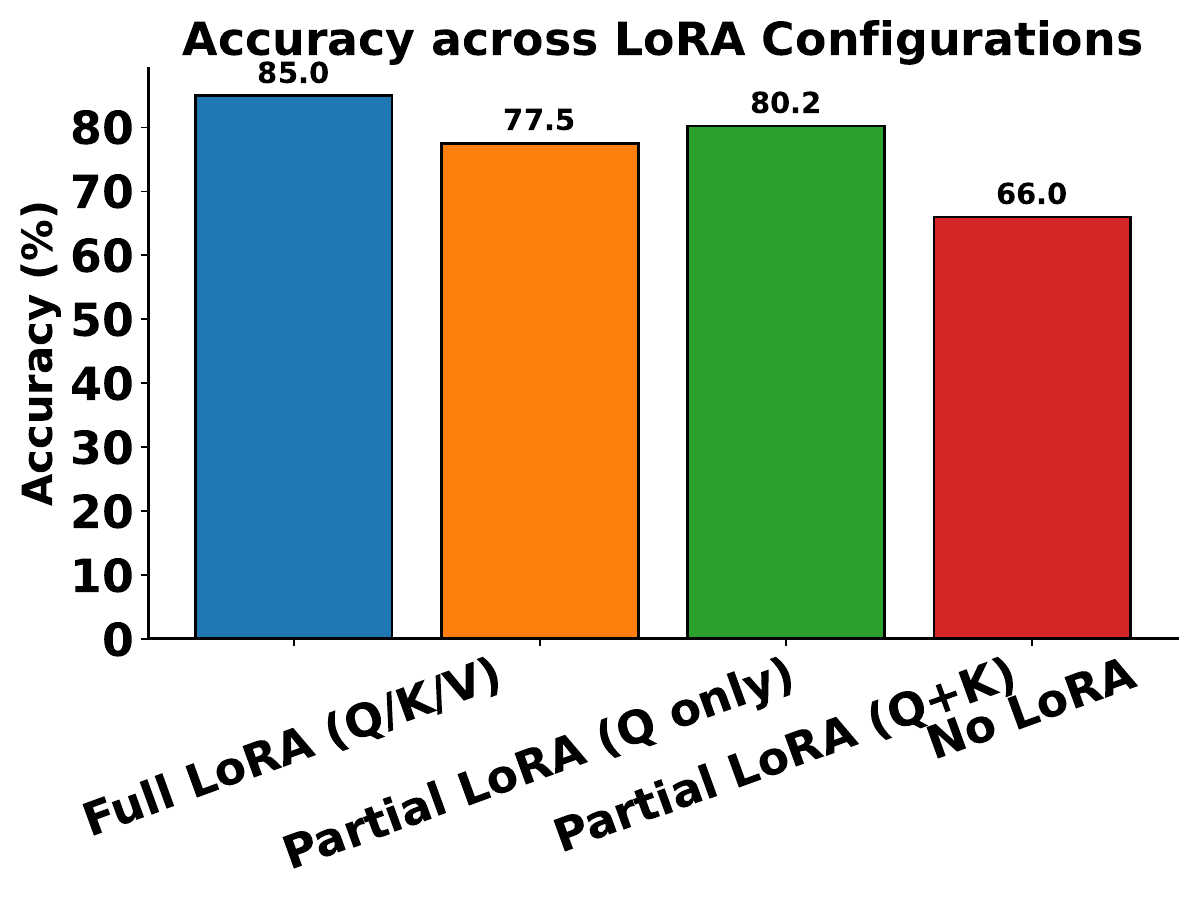}
        \caption{Accuracy across LoRA configurations.}
        \label{fig:lora_acc}
    \end{subfigure}
    \hfill
    \begin{subfigure}[b]{0.25\textwidth}
        \centering
        \includegraphics[width=\linewidth, trim=10 10 10 10, clip]{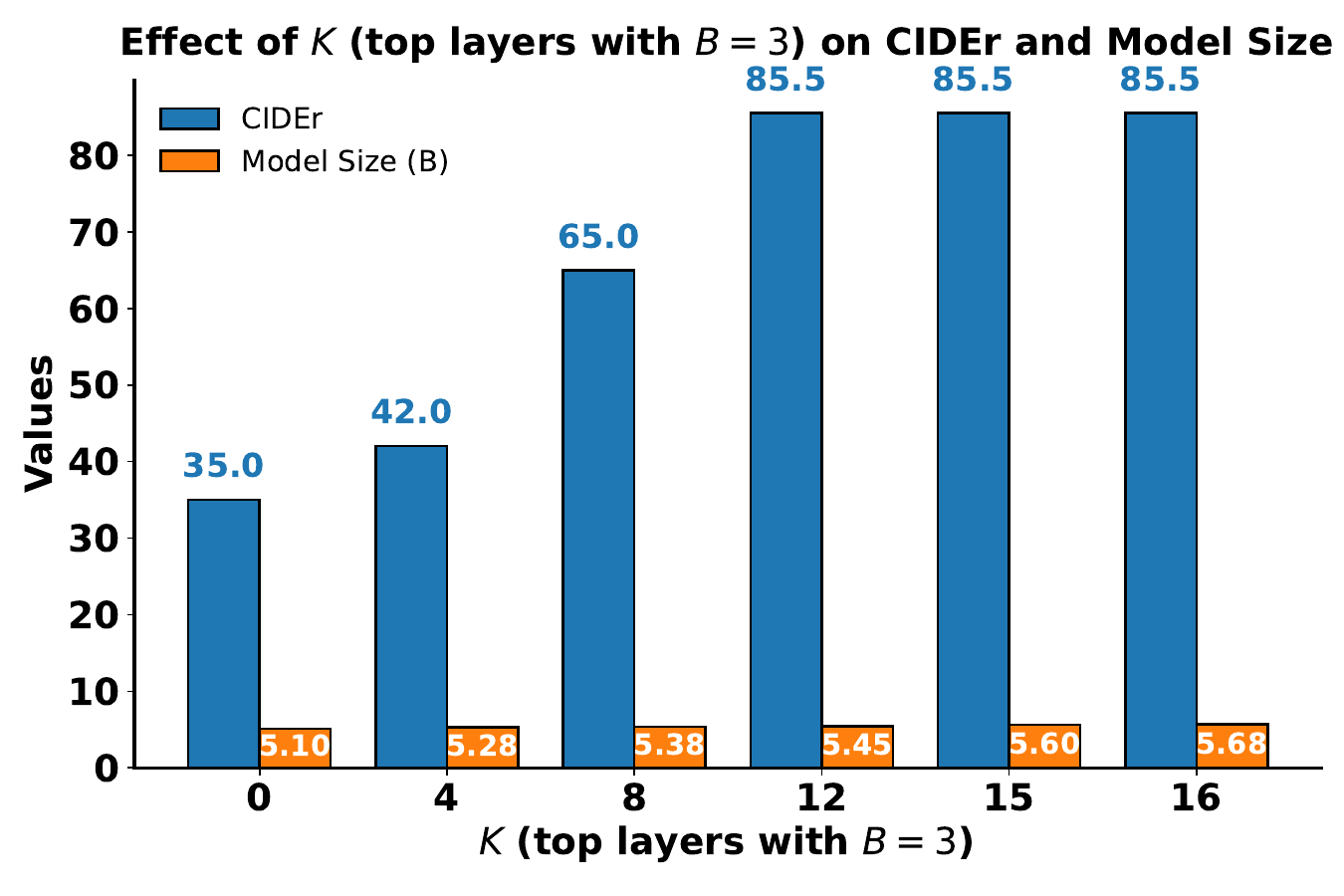}
        \caption{Effect of $K$ (top layers, $B_l{=}3$) on CIDEr and model size.}
        \label{fig:k_cider}
    \end{subfigure}

    \caption{Comprehensive experimental results:  
    \textbf{(a)} Increasing parameters improves CIDEr, from 59.7 at 4.35B to 88.9 at 8.29B.  
    \textbf{(b)} Higher $\alpha$ reduces validation CE, dropping from 15.0 at $\alpha{=}0.0$ to 1.05 at $\alpha{=}1.0$.  
    \textbf{(c)} LoRA placement impacts accuracy: full LoRA (Q/K/V) achieves 85.0\%, while no LoRA drops to 66.0\%.  
    \textbf{(d)} Raising $K$ (layers with $B_l{=}3$) boosts CIDEr from 35.0 at $K{=}0$ to 85.5 at $K{=}12$, with model size growing moderately from 5.10B to 5.68B. All the experiments are done on Qwen2.5-VL-7B.}
    \label{fig:all_results}
\end{figure*}

\noindent \textbf{Parameter Count vs. CIDEr (Figure~\ref{fig:param_cider}):}  
Increasing the parameter count leads to clear improvements in generation quality. The CIDEr score rises steadily from 59.7 at 4.35B parameters to 88.9 at 8.29B, highlighting the strong correlation between model scale and performance. This trend suggests that larger models are able to capture richer cross-modal dependencies, though at the cost of greater computational demand.  

\noindent \textbf{$\alpha$ vs. Validation CE (Figure~\ref{fig:alpha_valce}):}  
The effect of $\alpha$ on optimization is pronounced and consistent. Validation cross-entropy decreases smoothly from 15.0 at $\alpha{=}0.0$ to 1.05 at $\alpha{=}1.0$, demonstrating how higher $\alpha$ values improve convergence and reduce error. This steady reduction emphasizes the role of $\alpha$ as a balancing factor for stabilizing training dynamics and enhancing model reliability.  

\noindent \textbf{LoRA Configurations (Figure~\ref{fig:lora_acc}):} The comparison of LoRA variants shows how adaptation choices impact performance. Full LoRA applied to Q/K/V projections achieves the highest accuracy at 85.0\%, confirming the benefit of updating all key projections. Partial variants (77.5\% for Q-only and 80.2\% for Q+K) perform moderately well, while removing LoRA altogether drops accuracy sharply to 66.0\%. These results underline the importance of careful module selection in parameter-efficient tuning.  

\noindent\textbf{Effect of $K$ (Figure~\ref{fig:k_cider}):}
Increasing the number of top layers assigned $B{=}3$ yields consistent gains, with CIDEr rising from 35.0 at $K{=}0$ (all $B{=}2$) to 85.5 at $K{=}12$, while the parameter count grows only from 5.10B to 5.68B. This shows that selectively deepening higher layers is an efficient way to enhance expressiveness without heavy computational cost. Beyond $K{=}12$, additional increases in $K$ raise parameter count but deliver no meaningful improvements, indicating diminishing returns once sufficient upper-layer capacity is allocated. This pattern aligns with the sensitivity rationale: output-proximal layers exhibit larger Fisher traces and are therefore more susceptible to approximation error, making them the most beneficial layers to equip with higher PHM capacity.


\noindent\textbf{Transition Stabilization Components (Table~\ref{tab:stability_ablation}):}
We analyze the core mechanisms that stabilize the dense$\rightarrow$PHM transition: KD, reconstruction loss, and the residual interpolation path. Removing KD causes the largest drop, showing its importance for preserving logit behavior. Removing reconstruction also hurts convergence, as operator-level alignment guides PHM layers toward their dense counterparts. Disabling the residual path forces an abrupt substitution that destabilizes training. Uniform PHM settings (B=2 or B=3) remain inferior to our selective capacity assignment, confirming the benefit of allocating higher PHM capacity only where sensitivity is highest.

\subsection{FLOPs Reduction and Inference Efficiency}
\begin{table}[!t]
\centering
\setlength{\tabcolsep}{2pt} 
\caption{Parameter counts (Billion, B), FLOPs (Trillion, T), and average inference time (ms) for base models and our PHM-compressed variants. PHM delivers consistent computational savings across all architectures, with Qwen2.5-VL-7B exhibiting the largest gains in latency and FLOPs efficiency.}
\begin{tabular}{lccc}
\toprule
\small
\textbf{Model} 
& \textbf{Parameters} 
& \textbf{FLOPs} 
& \textbf{Inference Time} \\
& base $\rightarrow$ ours 
& base $\rightarrow$ ours 
& base $\rightarrow$ ours \\
\midrule

InstructBLIP
& 7B $\rightarrow$ \textbf{\textcolor{green}{5.45B}}
& 4.50 $\rightarrow$\textbf{\textcolor{green} {3.31}}
& 2.10 $\rightarrow$ \textcolor{green}{\textbf{1.20}} \\

LLaVA-1.5
& 7B $\rightarrow$ \textbf{\textcolor{green}{5.15B}}
& 4.70 $\rightarrow$ \textbf{\textcolor{green}{3.48}}
& 1.95 $\rightarrow$ \textcolor{green}{\textbf{1.05}} \\

Qwen2.5-VL
& 7B $\rightarrow$ \textbf{\textcolor{green}{5.38B}}
& 4.94 $\rightarrow$ \textbf{\textcolor{green}{3.64}}
& 1.70 $\rightarrow$ \textcolor{green}{\textbf{0.93}} \\

\bottomrule
\end{tabular}
\label{tab:flops_reduction}
\end{table}
Table~\ref{tab:flops_reduction} summarizes average forward FLOPs and inference times before and after PHM reparameterization. PHM consistently cuts computation by 25–30\% with notable latency gains: InstructBLIP drops from 4.50 to 3.31T FLOPs (2.10→1.20 ms), LLaVA-1.5-7B from 4.70 to 3.48T (1.95→1.05 ms), and Qwen2.5-VL-7B from 4.94 to 3.64T (1.70→0.93 ms), nearly halving runtime. Beyond reducing memory footprint, PHM delivers practical speedups without harming accuracy—and in some cases (e.g., Qwen2.5-VL-7B) even improves it, making it a strong accuracy–efficiency trade-off for large-scale deployment.

\begin{table}[!t]
\centering
\small
\setlength{\tabcolsep}{2.5pt} 
\caption{Ablation on the stabilization components used during the 
dense$\rightarrow$PHM transition. A green checkmark (\cmark) denotes that a component is enabled, while a red cross (\xmark) indicates its removal. Each row modifies one or more of the three stabilizing mechanisms: KD, reconstruction loss, and the residual interpolation path, or varies the PHM capacity ($B_l$). Results are reported on the Flickr30k 
validation set using Qwen2.5-VL-7B.}
\label{tab:stability_ablation}
\begin{tabular}{@{}lcccccc@{}}
\toprule
\textbf{Variant} & \textbf{KD} & \textbf{Recon} & \textbf{Residual} & 
\textbf{CIDEr} & \textbf{BLEU-4} & \textbf{Params} \\
\midrule
Full (Ours)            & \cmark & \cmark & \cmark & \textbf{85.2} & \textbf{31.0} & \textbf{5.38B} \\
Variant A                        & \xmark & \cmark & \cmark & 26.5 & 4.2  & 5.38B \\
Variant B            & \cmark & \xmark & \cmark & 46.7 & 9.5  & 5.38B \\
Variant C       & \cmark & \cmark & \xmark & 57.6 & 3.4  & 5.38B \\
PHM ($B_l$=2)       & \cmark & \cmark & \cmark & 65.9 & 21.3 & 4.11B \\
PHM ($B_l$=3)       & \cmark & \cmark & \cmark & 84.7 & 29.4 & 6.65B \\
\bottomrule
\end{tabular}
\end{table}

\section{Conclusion}
This work demonstrates the effectiveness of unifying parameter-efficient tuning, hypercomplex parameterization, and knowledge distillation for scalable multimodal transformers. Our approach directly reduces parameter size, FLOPs, and inference time while preserving semantic fidelity, achieving consistent efficiency–performance gains on captioning and question answering tasks. Residual hypercomplex blocks with convex blending and reconstruction loss stabilized dense–hypercomplex transitions, while a dense-teacher vs. hypercomplex-student KD strategy ensured semantic alignment. Together, these advances provide a principled path toward efficient multimodal models.  

\noindent \textbf{Future works:} Future directions include extending evaluation to audio and video modalities, refining hypercomplex capacity allocation, and integrating quantization-aware training for deployment in resource-constrained settings, with potential applications in robotics, healthcare, edge AI.

{
    \small
    \bibliographystyle{unsrt}
    \bibliography{main}

@String(CVPR= {IEEE Conf. Comput. Vis. Pattern Recog.})

@String(ICCV= {Int. Conf. Comput. Vis.})

@String(ECCV= {Eur. Conf. Comput. Vis.})

@String(NIPS= {Adv. Neural Inform. Process. Syst.})

@String(ICLR = {Int. Conf. Learn. Represent.})

@String(AAAI = {AAAI})

@String(CVPR  = {CVPR})

@String(ICCV  = {ICCV})

@String(ECCV  = {ECCV})

@String(NIPS  = {NeurIPS})

@String(ICLR  = {ICLR})

@article{Gadre2023,
  title={CLIP meets multimodality: A survey of vision-and-language models},
  author={Gadre, Shubham and Yang, Chenguang and Mao, Jiahui and Bisk, Yonatan and Chai, Joyce Y.},
  journal={Transactions of the Association for Computational Linguistics},
  volume={11},
  pages={1550--1568},
  year={2023}
}

@article{OpenAI2023GPT4,
  title={GPT-4 Technical Report},
  author={{OpenAI}},
  journal={arXiv preprint arXiv:2303.08774},
  year={2023}
}

@article{Zhao2023Survey,
  title={A Comprehensive Survey of Multimodal Large Language Models},
  author={Zhao, Wayne Xin and Zhou, Kun and Li, Junyi and Tang, Tianyi and Wang, Xiaoxuan and Hou, Yifan and Chen, Zican and Zhang, Jing and Wang, Kai and Li, Weizhi and others},
  journal={arXiv preprint arXiv:2306.13549},
  year={2023}
}

@inproceedings{Tsimpoukelli2021,
  title={Multimodal Few-Shot Learning with Frozen Language Models},
  author={Tsimpoukelli, Maria and Menick, Jacob and Cabi, Serkan and Eslami, S. M. Ali and Vinyals, Oriol and Hill, Felix},
  booktitle={NeurIPS},
  year={2021}
}

@inproceedings{Zeng2023,
  title={GLM-130B: An Open Bilingual Pre-trained Model},
  author={Zeng, Aohan and Liu, Xiao and Du, Zhengxiao and Wang, Zihan and others},
  booktitle={ICLR},
  year={2023}
}

@article{Xu2022,
  title={A Survey on Multimodal Pre-trained Models},
  author={Xu, Han and Sun, Wei and Li, Hongzhi and others},
  journal={IEEE Transactions on Knowledge and Data Engineering},
  year={2022}
}

@inproceedings{Vasudevan2023,
  title={Foundational Models in Robotics: Applications, Challenges, and Opportunities},
  author={Vasudevan, Arun and Chen, Xi and Gupta, Abhinav},
  booktitle={RSS},
  year={2023}
}

@inproceedings{Hu2021LoRA,
  title={LoRA: Low-Rank Adaptation of Large Language Models},
  author={Hu, Edward and Shen, Yelong and Wallis, Phillip and others},
  booktitle={ICLR},
  year={2021}
}

@inproceedings{Mahabadi2021Compacter,
  title={Compacter: Efficient Low-Rank Hypercomplex Adapter Layers},
  author={Mahabadi, Rabeeh Karimi and Henderson, James and Ruder, Sebastian},
  booktitle={NeurIPS},
  year={2021}
}

@inproceedings{He2022UnifiedPETuning,
  title={Towards Unified Parameter-Efficient Tuning},
  author={He, Pengcheng and Gao, Jianfeng and Chen, Weizhu},
  booktitle={ACL},
  year={2022}
}

@inproceedings{Dettmers2023QLoRA,
  title={QLoRA: Efficient Finetuning of Quantized LLMs},
  author={Dettmers, Tim and Lewis, Mike and Shleifer, Sam and Zettlemoyer, Luke},
  booktitle={NeurIPS},
  year={2023}
}

@inproceedings{Asai2022Multitask,
  title={Multitask Instruction Tuning of Pretrained Transformers},
  author={Asai, Akari and Hajishirzi, Hannaneh},
  booktitle={EMNLP},
  year={2022}
}

@article{Jin2023SurveyKD,
  title={Knowledge Distillation: A Comprehensive Survey},
  author={Jin, Xiaoyan and Wang, Dong and others},
  journal={IEEE Transactions on Pattern Analysis and Machine Intelligence},
  year={2023}
}

@article{Hsieh2023DistillLM,
  title={Distilling Large Language Models for Efficient Deployment},
  author={Hsieh, Cheng-Yu and Li, Rui and Yu, Dong},
  journal={Foundations and Trends in Machine Learning},
  year={2023}
}

@inproceedings{Zhou2023MMKD,
  title={Multimodal Knowledge Distillation for Efficient Multimodal Transformers},
  author={Zhou, Liang and Xu, Yichong and Wang, Xin},
  booktitle={ACL},
  year={2023}
}

@inproceedings{Fang2023CrossModal,
  title={Cross-Modal Distillation with Semantic Alignment},
  author={Fang, Hao and Zhou, Lei and Zhang, Yu},
  booktitle={ICCV},
  year={2023}
}

@inproceedings{Fedus2022Switch,
  title     = {Switch Transformers: Scaling to Trillion Parameter Models with Simple and Efficient Sparsity},
  author    = {William Fedus and Barret Zoph and Noam Shazeer},
  booktitle = {Proceedings of the International Conference on Learning Representations (ICLR)},
  year      = {2022}
}

@inproceedings{Clark2022UnifiedSparsity,
  title     = {Unified Scaling Laws for Sparsity and Quantization in Neural Networks},
  author    = {Jonathan Clark and Colin Raffel and Anselm Levskaya and Noam Shazeer and Luke Metz},
  booktitle = {Proceedings of the International Conference on Machine Learning (ICML)},
  year      = {2022}
}

@inproceedings{Artetxe2021EfficientLMs,
  title     = {Efficient Large Scale Language Modeling with Mixtures of Experts},
  author    = {Mikel Artetxe and Shruti Bhosale and Naman Goyal and Luke Zettlemoyer},
  booktitle = {Proceedings of the Conference on Empirical Methods in Natural Language Processing (EMNLP)},
  year      = {2021}
}

@inproceedings{Parcollet18,
  title={Quaternion Recurrent Neural Networks},
  author={Parcollet, Titouan and Morchid, Mohamed and Linarès, Georges},
  booktitle={ICLR},
  year={2018}
}

@inproceedings{Gaudet18,
  title={Deep Quaternion Networks},
  author={Gaudet, Chase J. and Maida, Anthony S.},
  booktitle={IJCNN},
  year={2018}
}

@article{Zhu20,
  title={Octonion Recurrent Neural Networks},
  author={Zhu, Ying and Xu, Yong and Zhang, Peng},
  journal={Neurocomputing},
  volume={402},
  pages={148--160},
  year={2020}
}

@article{Zhang21Survey,
  title={Survey on Quaternion and Hypercomplex Neural Networks},
  author={Zhang, Shilin and Qi, Guo-Jun},
  journal={Artificial Intelligence Review},
  volume={54},
  number={5},
  pages={3781--3816},
  year={2021}
}

@inproceedings{Zhang21PHM,
  title={Parameterized Hypercomplex Multiplication Layers},
  author={Zhang, Shilin and Tay, Yi and Luu, Anh Tuan and Fu, Jie},
  booktitle={ICML},
  year={2021}
}

@inproceedings{Mao23,
  title={PHM-Transformers: Parameterized Hypercomplex Multiplication for Transformers},
  author={Mao, Rui and others},
  booktitle={TMLR},
  year={2023}
}

@inproceedings{Gao23,
  title={Hypercomplex Vision Transformers},
  author={Gao, Wei and Chen, Wei and Zhao, Xing},
  booktitle={CVPR},
  year={2023}
}

@article{Cheng24,
  title={Residual Hypercomplex Networks for Stable Deep Training},
  author={Cheng, Li and Liu, Hao},
  journal={Neural Networks},
  year={2024}
}

@article{Wang22,
  title={Hypercomplex Neural Networks with Enhanced Residual Connections},
  author={Wang, Li and Zhang, Hao and Xu, Kai},
  journal={IEEE Transactions on Neural Networks and Learning Systems},
  volume={33},
  number={12},
  pages={12345--12357},
  year={2022}
}

@inproceedings{Kim23,
  title={Hybrid Hypercomplex Transformers for Multimodal Learning},
  author={Kim, Jiyoung and Lee, Sungho and Park, Min},
  booktitle={AAAI},
  year={2023}
}

@article{Zhang24HCNN,
  title={Hypercomplex Neural Architectures for Large-Scale Multimodal Learning},
  author={Zhang, Rui and Han, Yujia},
  journal={IEEE Transactions on Pattern Analysis and Machine Intelligence},
  year={2024}
}

@inproceedings{Lee24Hybrid,
  title={Hybrid Hypercomplex Models for Cross-Modal Alignment},
  author={Lee, Minho and Park, Soojin},
  booktitle={ACL},
  year={2024}
}

@inproceedings{Hinton15,
  title={Distilling the Knowledge in a Neural Network},
  author={Hinton, Geoffrey and Vinyals, Oriol and Dean, Jeff},
  booktitle={NIPS Workshop},
  year={2015}
}

@inproceedings{Romero15,
  title={FitNets: Hints for Thin Deep Nets},
  author={Romero, Adriana and Ballas, Nicolas and Kahou, Samira Ebrahimi and others},
  booktitle={ICLR},
  year={2015}
}

@inproceedings{Zagoruyko17,
  title={Paying More Attention to Attention},
  author={Zagoruyko, Sergey and Komodakis, Nikos},
  booktitle={ICLR},
  year={2017}
}

@inproceedings{Park19,
  title={Relational Knowledge Distillation},
  author={Park, Wonpyo and Kim, Dongju and Lu, Yan and Cho, Minsu},
  booktitle={CVPR},
  year={2019}
}

@inproceedings{Sun19,
  title={Patient Knowledge Distillation for BERT Model Compression},
  author={Sun, Siqi and Cheng, Yu and Gan, Zhe and Liu, Jingjing},
  booktitle={EMNLP},
  year={2019}
}

@inproceedings{Jiao20,
  title={TinyBERT: Distilling BERT for Natural Language Understanding},
  author={Jiao, Xiaoqi and Yin, Yichun and Shang, Lifeng and others},
  booktitle={Findings of ACL},
  year={2020}
}

@inproceedings{Sanh19,
  title={DistilBERT: Smaller, Faster, Cheaper and Lighter},
  author={Sanh, Victor and Debut, Lysandre and Chaumond, Julien and Wolf, Thomas},
  booktitle={NeurIPS Workshop},
  year={2019}
}

@article{Gupta22,
  title={Multimodal Knowledge Distillation for Visual Question Answering},
  author={Gupta, Ankit and Sharma, Akshay and others},
  journal={IEEE Transactions on Multimedia},
  year={2022}
}

@inproceedings{Li23,
  title={Efficient Multimodal Knowledge Distillation},
  author={Li, Ming and Zhou, Chen and Wu, Yang},
  booktitle={AAAI},
  year={2023}
}

@article{Wu24,
  title={Cross-Modal Knowledge Distillation with Adaptive Teachers},
  author={Wu, Chen and Li, Zihan},
  journal={Pattern Recognition},
  year={2024}
}

@inproceedings{Mirzadeh20,
  title={Improved Knowledge Distillation via Teacher Assistant},
  author={Mirzadeh, Seyed Iman and Farajtabar, Mehrdad and Li, Ang and others},
  booktitle={AAAI},
  year={2020}
}

@inproceedings{Shen20,
  title={Progressive Knowledge Distillation for Efficient Visual Recognition},
  author={Shen, Zhihua and Savvides, Marios},
  booktitle={ECCV},
  year={2020}
}

@inproceedings{You17,
  title={Learning from Multiple Teachers},
  author={You, Shan and Xu, Chang and Xu, Chao and Tao, Dacheng},
  booktitle={ICML},
  year={2017}
}

@article{Yang21,
  title={Multi-Teacher Knowledge Distillation with Meta Learning},
  author={Yang, Wei and Xu, Jia and Wu, Ji},
  journal={IEEE Transactions on Neural Networks and Learning Systems},
  year={2021}
}

@inproceedings{Tang23,
  title={Cross-Modal Knowledge Distillation for Vision-Language Pretraining},
  author={Tang, Ruixiang and Chen, Xiang and Li, Chenliang},
  booktitle={ACL},
  year={2023}
}

@inproceedings{Wang23RLKD,
  title={Reinforcement Learning with Knowledge Distillation},
  author={Wang, Jiarui and Liu, Xinyu},
  booktitle={ICLR},
  year={2023}
}

@inproceedings{Chen24XKD,
  title={Cross-Modal Knowledge Distillation for Retrieval and Pretraining},
  author={Chen, Hao and Zhang, Yue},
  booktitle={ACL},
  year={2024}
}

@inproceedings{Houlsby19,
  title={Parameter-Efficient Transfer Learning for NLP},
  author={Houlsby, Neil and Giurgiu, Andrei and Jastrzebski, Stanislaw and others},
  booktitle={ICML},
  year={2019}
}

@inproceedings{Hu21,
  title={LoRA: Low-Rank Adaptation of Large Language Models},
  author={Hu, Edward and Shen, Yelong and Wallis, Phillip and others},
  booktitle={ICLR},
  year={2021}
}

@inproceedings{Zhang22AdaLoRA,
  title={Adaptive Low-Rank Adaptation for Parameter-Efficient Fine-Tuning},
  author={Zhang, Renjie and Zhou, Zhiyuan and Wang, Yu},
  booktitle={ICLR},
  year={2022}
}

@inproceedings{Liu23DoRA,
  title={DoRA: Weight-Decomposed Low-Rank Adaptation},
  author={Liu, Yuhan and Kim, Taewoon and Chen, Qian},
  booktitle={NeurIPS},
  year={2023}
}

@inproceedings{Chen24MixLoRA,
  title={MixLoRA: Composable Parameter-Efficient Adaptation via Mixtures of Low-Rank Experts},
  author={Chen, Mingyu and Li, Jiahao and Wu, Ziyang},
  booktitle={ACL},
  year={2024}
}

@inproceedings{Li21,
  title={Prefix-Tuning: Optimizing Continuous Prompts for Generation},
  author={Li, Xiang Lisa and Liang, Percy},
  booktitle={ACL},
  year={2021}
}

@inproceedings{Lester21,
  title={The Power of Scale: Parameter-Efficient Adaptation for Pretrained Language Models},
  author={Lester, Brian and Al-Rfou, Rami and Constant, Noah},
  booktitle={EMNLP},
  year={2021}
}

@inproceedings{Dettmers24,
  title={QLoRA: Efficient Finetuning of Quantized LLMs},
  author={Dettmers, Tim and Lewis, Mike and Shleifer, Sam and Zettlemoyer, Luke},
  booktitle={NeurIPS},
  year={2024}
}

@article{Zhang23,
  title={Multimodal LoRA for Vision-Language Models},
  author={Zhang, Yong and Liu, Haotian and others},
  journal={arXiv preprint arXiv:2305.15007},
  year={2023}
}

@inproceedings{Chen23,
  title={Parameter-Efficient Multimodal Learning with Low-Rank Adaptation},
  author={Chen, Zhe and Huang, Shuo and Wu, Xu},
  booktitle={NeurIPS},
  year={2023}
}

@article{Mukherjee23,
  title={PEFT for Multimodal Transformers},
  author={Mukherjee, Arijit and Sharma, Aditya},
  journal={Transactions on Machine Learning Research},
  year={2023}
}

@inproceedings{He24Unified,
  title={Unified Parameter-Efficient Fine-Tuning for Instruction-Tuned Models},
  author={He, Junxian and Wang, Yuchen and Zhou, Chunting and Neubig, Graham},
  booktitle={ICLR},
  year={2024}
}

@inproceedings{Ma24,
  title={Sparse-Tuning: Parameter-Efficient Fine-Tuning with Sparse Updates},
  author={Ma, Hao and Xu, Wei and Zhang, Peng},
  booktitle={ACL},
  year={2024}
}

@inproceedings{agrawal2019nocaps,
  title={Nocaps: Novel object captioning at scale},
  author={Agrawal, Harsh and Desai, Karan and Wang, Yufei and Chen, Xinlei and Jain, Rishabh and Johnson, Mark and Batra, Dhruv and Parikh, Devi and Lee, Stefan and Anderson, Peter},
  booktitle={Proceedings of the IEEE/CVF international conference on computer vision},
  pages={8948--8957},
  year={2019}
}

@inproceedings{lin2014microsoft,
  title={Microsoft coco: Common objects in context},
  author={Lin, Tsung-Yi and Maire, Michael and Belongie, Serge and Hays, James and Perona, Pietro and Ramanan, Deva and Doll{\'a}r, Piotr and Zitnick, C Lawrence},
  booktitle={European conference on computer vision},
  pages={740--755},
  year={2014},
  organization={Springer}
}

@inproceedings{plummer2015flickr30k,
  title={Flickr30k entities: Collecting region-to-phrase correspondences for richer image-to-sentence models},
  author={Plummer, Bryan A and Wang, Liwei and Cervantes, Chris M and Caicedo, Juan C and Hockenmaier, Julia and Lazebnik, Svetlana},
  booktitle={Proceedings of the IEEE international conference on computer vision},
  pages={2641--2649},
  year={2015}
}

@article{lu2022learn,
  title={Learn to explain: Multimodal reasoning via thought chains for science question answering},
  author={Lu, Pan and Mishra, Swaroop and Xia, Tanglin and Qiu, Liang and Chang, Kai-Wei and Zhu, Song-Chun and Tafjord, Oyvind and Clark, Peter and Kalyan, Ashwin},
  journal={Advances in Neural Information Processing Systems},
  volume={35},
  pages={2507--2521},
  year={2022}
}

@article{luo2025finmme,
  title={FinMME: Benchmark Dataset for Financial Multi-Modal Reasoning Evaluation},
  author={Luo, Junyu and Kou, Zhizhuo and Yang, Liming and Luo, Xiao and Huang, Jinsheng and Xiao, Zhiping and Peng, Jingshu and Liu, Chengzhong and Ji, Jiaming and Liu, Xuanzhe and others},
  journal={arXiv preprint arXiv:2505.24714},
  year={2025}
}

@article{bai2025qwen2,
  title={Qwen2. 5-vl technical report},
  author={Bai, Shuai and Chen, Keqin and Liu, Xuejing and Wang, Jialin and Ge, Wenbin and Song, Sibo and Dang, Kai and Wang, Peng and Wang, Shijie and Tang, Jun and others},
  journal={arXiv preprint arXiv:2502.13923},
  year={2025}
}

@article{dai2023instructblip,
  title={Instructblip: Towards general-purpose vision-language models with instruction tuning},
  author={Dai, Wenliang and Li, Junnan and Li, Dongxu and Tiong, Anthony and Zhao, Junqi and Wang, Weisheng and Li, Boyang and Fung, Pascale N and Hoi, Steven},
  journal={Advances in neural information processing systems},
  volume={36},
  pages={49250--49267},
  year={2023}
}

@inproceedings{liu2024improved,
  title={Improved baselines with visual instruction tuning},
  author={Liu, Haotian and Li, Chunyuan and Li, Yuheng and Lee, Yong Jae},
  booktitle={Proceedings of the IEEE/CVF conference on computer vision and pattern recognition},
  pages={26296--26306},
  year={2024}
}

@article{liu2024llava,
  title={Llava-next: Improved reasoning, ocr, and world knowledge, January 2024},
  author={Liu, Haotian and Li, Chunyuan and Li, Yuheng and Li, Bo and Zhang, Yuanhan and Shen, Sheng and Lee, Yong Jae},
  journal={URL https://llava-vl. github. io/blog/2024-01-30-llava-next},
  volume={1},
  number={8},
  year={2024}
}

@article{wu2024deepseek,
  title={Deepseek-vl2: Mixture-of-experts vision-language models for advanced multimodal understanding},
  author={Wu, Zhiyu and Chen, Xiaokang and Pan, Zizheng and Liu, Xingchao and Liu, Wen and Dai, Damai and Gao, Huazuo and Ma, Yiyang and Wu, Chengyue and Wang, Bingxuan and others},
  journal={arXiv preprint arXiv:2412.10302},
  year={2024}
}

@article{hu2023large,
  title={Large multilingual models pivot zero-shot multimodal learning across languages},
  author={Hu, Jinyi and Yao, Yuan and Wang, Chongyi and Wang, Shan and Pan, Yinxu and Chen, Qianyu and Yu, Tianyu and Wu, Hanghao and Zhao, Yue and Zhang, Haoye and others},
  journal={arXiv preprint arXiv:2308.12038},
  year={2023}
}

@inproceedings{yu2025mquant,
  title={Mquant: Unleashing the inference potential of multimodal large language models via static quantization},
  author={Yu, JiangYong and Zhou, Sifan and Yang, Dawei and Li, Shuoyu and Wang, Shuo and Hu, Xing and Xu, Chen and Xu, Zukang and Shu, Changyong and Yuan, Zhihang},
  booktitle={Proceedings of the 33rd ACM International Conference on Multimedia},
  pages={1783--1792},
  year={2025}
}

@inproceedings{li2025mbq,
  title={Mbq: Modality-balanced quantization for large vision-language models},
  author={Li, Shiyao and Hu, Yingchun and Ning, Xuefei and Liu, Xihui and Hong, Ke and Jia, Xiaotao and Li, Xiuhong and Yan, Yaqi and Ran, Pei and Dai, Guohao and others},
  booktitle={Proceedings of the Computer Vision and Pattern Recognition Conference},
  pages={4167--4177},
  year={2025}
}
}

\end{document}